\documentclass[letterpaper]{article} 
\usepackage{aaai2027}  

\usepackage[hyphens]{url}  
\usepackage{graphicx} 
\usepackage{natbib}  
\usepackage{caption} 
\usepackage{algorithm}
\usepackage{algorithmic}

\usepackage{newfloat}
\usepackage{listings}
\DeclareCaptionStyle{ruled}{labelfont=normalfont,labelsep=colon,strut=off} 
\floatstyle{ruled}
\newfloat{listing}{tb}{lst}{}
\floatname{listing}{Listing}

\usepackage{booktabs}
\usepackage{amsmath}
\usepackage[table]{xcolor}

\title{Memory as Belief: Bayesian Evolution of Memory for Self-Evolving LLM Agents}
\author{
    Yiming Xu\textsuperscript{\rm 1,2},
    Qian Yang\textsuperscript{\rm 1,2},
    Zihan Chen\textsuperscript{\rm 3},
    Yuefei Wu\textsuperscript{\rm 4},
    Wenhan Zeng\textsuperscript{\rm 1},
    Bin Shi\textsuperscript{\rm 1,2},
    Bo Dong\textsuperscript{\rm 1,2}
}
\affiliations{
    \textsuperscript{\rm 1}School of Computer Science and Technology, Xi'an Jiaotong University\\
    \textsuperscript{\rm 2}Shaanxi Provincial Key Laboratory of Big Data Knowledge Engineering, Xi’an Jiaotong University\\
    \textsuperscript{\rm 3}University of Virginia, Charlottesville, Virginia, USA\\
    \textsuperscript{\rm 4} School of Cyber Science and Engineering, Zhengzhou University\\
    
    \{xuyiming, shibin, dong.bo\}@xjtu.edu.cn, yangqian@stu.xjtu.edu.cn,brf3rx@virginia.edu,wuyfei@zzu.edu.cn
}

\begin{document}

\maketitle

\begin{abstract}
Large language model (LLM) agents increasingly rely on external memory to accumulate past experiences for continual adaptation. However, effectively reusing memory requires accurately assessing its value, as retrieved experiences may become unreliable when transferred across task contexts. Despite its importance, memory valuation remains largely underexplored, with existing methods relying on historical retrieval success rates that overlook the context dependency of memory usefulness and the varying reliability of estimates under different amounts of evidence. 
Inspired by the Bayesian brain hypothesis, we propose BEAM (Bayesian-style Evolutionary Agentic Memory), a framework that reformulates memory value evaluation as a parameterized cognitive inference process, where memory value is represented as a context-conditioned dynamic belief rather than a passive historical statistic. Specifically, BEAM infers memory value beliefs from current task contexts and continuously calibrates them using execution feedback as empirical evidence, transforming memory from passive storage into an adaptive cognitive mechanism for continual self-evolution.
Extensive experiments across reasoning and code generation benchmarks demonstrate that BEAM consistently improves memory utilization and enables continual self-evolution of LLM agents.

\end{abstract}

\section{Introduction}

Large language models (LLMs) have demonstrated remarkable efficacy in natural language understanding, code generation, knowledge-intensive QA, and complex reasoning~\citep{yao2023tree,zhao2024expel,jimenez2024swe,singh2025openai}. As LLMs increasingly serve as the foundation for autonomous agents, the deployment paradigm is shifting from static benchmark evaluations to continuous streams of open-world tasks~\citep{xie2023openagents,xu2026theagentcompany}.
However, when operating over continuous task streams, contemporary agent systems largely fail to learn from accumulated cross-task experience~\citep{zhang2026g,xu2026mem}. Since they process each task in isolation, agents repeatedly encounter similar failures and overlook valuable transferable insights from related tasks~\citep{ouyang2026reasoningbank}. Therefore, enabling agents to accumulate and leverage valuable experiences across tasks has become a fundamental challenge toward achieving reliable self-evolving systems.

In practical deployment, continuously fine-tuning or online-updating model parameters could be computationally expensive and engineering-intensive~\citep{zhao2023does,xia2025experience}. External memory modules have therefore been widely adopted to store experience and knowledge that can be reused across tasks~\citep{hu2025memory}. 
However, effectively leveraging external memories remains challenging, as stored experiences are not always reliable and may become ineffective or even misleading when transferred to new task contexts~\citep{ouyang2026reasoningbank}. Existing studies~\citep{xu2026mem,zhang2026g,han2025legomem} rarely provide a principled assessment of whether a retrieved memory remains valuable under the current task context. 
Recently, a few studies predominantly estimate memory value through heuristic scoring mechanisms, such as historical retrieval success rates or manually designed importance scores~\citep{zhang2026memrl,xia2025experience,zhong2024memorybank}. However, these metrics implicitly treat memory value as a task-independent and scalar quantity. In reality, memory value is inherently context-dependent and subject to uncertainty based on accumulated evidence. For instance, a reasoning memory containing a \textit{"depth-first recursive traversal"} strategy may achieve a high success rate on small-scale algorithmic tasks, but fails when transferred to large-scale problems with strict computational or stack constraints. Moreover, scalar metrics fail to distinguish a memory proven over hundreds of trials from one that succeeded merely once by chance. This raises a central question: \textit{How can agents dynamically evaluate and calibrate context-dependent memory values under uncertainty, thereby enabling reliable and continual evolution rather than merely accumulating experiences?}

To overcome these limitations, we revisit memory value estimation from the perspective of the Bayesian brain hypothesis~\citep{knill2004bayesian}. This hypothesis posits that cognitive systems continuously update their probabilistic beliefs about the world by integrating prior experience with incoming observations~\citep{friston2010free,pouget2013probabilistic}. 
Inspired by this principle, our key insight is to formulate memory value evaluation as a parameterized cognitive inference process, where memory value is represented as a context-conditioned dynamic belief rather than a passive historical statistic. From this perspective, the parameterized cognitive system first infers the prior value of memories based on the current task context. After a memory is involved in reasoning and decision-making, the resulting execution feedback serves as empirical evidence for updating the posterior belief over its value. This belief calibration process transforms memory from passive storage into an adaptive cognitive mechanism, enabling LLM agents to continually refine their experience and self-evolve under open-world task streams.

In this work, we propose BEAM (Bayesian-style Evolutionary Agentic Memory), a framework that enables LLM agents to perform continual memory valuation through context-conditioned belief modeling and evidence-driven calibration. The proposed framework consists of three components. First, we construct structured reasoning memories from historical trajectories by distilling transferable strategies while filtering task-specific details. Second, we design a context-conditioned memory belief model, which is a lightweight LoRA-adapted estimator that carries out this cognitive inference without modifying the backbone LLM. Given the task context and candidate memories, it predicts prior belief distributions over memory utility, enabling uncertainty-aware memory valuation and subsequent belief calibration. Third, we introduce an evidence-driven belief updating mechanism that leverages trajectory execution outcomes as empirical evidence to obtain memory-level supervision and performs evidential calibration, preference-aware optimization, and selective regularization for continual value refinement. Extensive experiments across diverse reasoning and code generation benchmarks demonstrate that BEAM significantly enhances memory utilization and drives continual self-evolution in LLM agents. Our contributions are summarized as follows:

$\bullet$ \textbf{\textit{Perspective.}} We revisit memory valuation through the lens of the Bayesian brain hypothesis, transforming memory from a passive storage structure into an adaptive cognitive mechanism capable of dynamic belief calibration.

$\bullet$ \textbf{\textit{Algorithm.}} We propose BEAM, a Bayesian memory evolution framework that integrates reasoning memory construction, context-conditioned belief modeling, and evidence-driven belief calibration for continual memory valuation.

$\bullet$ \textbf{\textit{Experiments.}} Extensive experiments across diverse reasoning and code generation benchmarks demonstrate that BEAM consistently outperforms state-of-the-art baselines in memory utilization and drives long-term continual self-evolution for LLM agents.

\section{Related Work}

\textbf{Large Language Model Agents.} 
Large language model (LLM) agents extend the capabilities of language models on complex tasks by integrating task planning, tool use, and environment interaction~\citep{wang2024survey}. To further enhance agent capabilities, existing studies have explored reasoning-action coordination, reflective refinement, and experience reuse~\citep{zhang2025landscape}. ReAct~\citep{yao2022react} interleaves reasoning and acting, allowing models to adapt subsequent actions based on observations; Reflexion~\citep{shinn2023reflexion} uses verbal feedback to revise future decisions; Voyager~\citep{wang2023voyager} continuously explores open-ended environments and accumulates reusable skills; and Synapse~\citep{zheng2024synapse} retrieves historical trajectory exemplars to support long-horizon task execution. These studies demonstrate that feedback and accumulated experience can enhance agents' adaptability across tasks. However, many existing agents still rely heavily on current-context prompting, heuristic reflection, or limited historical examples for capability improvement, lacking stable mechanisms for experience consolidation, value assessment, and online updating~\citep{gao2025survey}. How to accumulate reusable experience and continuously optimize its selection and use remains a key challenge for self-evolving agents. 

\noindent\textbf{Agent Memory Organization.}
External memory allows agents to continually accumulate, organize and reuse interaction experience without modifying the backbone model, making it an important mechanism for long-term adaptation~\citep{wang2025scm}. Existing studies have explored diverse forms of memory representation and organization. MemoryBank~\citep{zhong2024memorybank} and MemGPT~\citep{packer2023memgpt} focus on long-term user memory and hierarchical context management; ReasoningBank~\citep{ouyang2026reasoningbank} distills successful and failed trajectories into reusable reasoning memories; A-Mem~\citep{xu2026mem}, Zep~\citep{rasmussen2025zep}, and G-Memory~\citep{zhang2026g} organize memories using graph-based or hierarchical structures; and LEGOMem~\citep{han2025legomem} and Buffer of Thoughts~\citep{yang2024buffer} abstract experience into procedural modules or thought templates to guide subsequent task solving. These efforts enable agents to accumulate and structure past experience~\citep{hu2025memory}. However, stored memories may contain low-value or noisy experiences, leading to suboptimal reasoning and decision-making~\citep{wei2025evo}.

\noindent\textbf{Agent Memory Value Evaluation.}
Since stored memories are not necessarily reliable or beneficial, evaluating their value has increasingly attracted attention~\citep{yu2026agentic}. Existing studies~\citep{zhong2024memorybank} primarily explore heuristic-driven approaches to model and quantify memory utility, such as historical retrieval success rates. MemRL~\citep{zhang2026memrl} estimates memory benefits using Q-values; From Experience to Strategy~\citep{xia2025experience} updates strategy weights based on counterfactual reward differences; and PlugMem~\citep{yang2026plugmem} quantifies the decision-relevant information gain per memory token from a utility--cost perspective. Another line of work learns memory operations or update policies: Memory-R1~\citep{yan2026memory} and Mem-$\alpha$~\citep{wang2025mem} learn strategies for memory construction, management, or updating, while MemSkill~\citep{zhang2026memskill} and ReMe~\citep{cao2026remember} further explore the dynamic evolution of memory skills and procedural experience. Although recent studies have advanced memory utility estimation and management, they typically model memory value as a static scalar, overlooking its uncertainty and context-dependent reliability~\citep{zhao2024expel}. In contrast, we model memory value as a context-conditioned dynamic belief that is continuously refined through online feedback.

\begin{figure*}
  \centering
  \includegraphics[width=0.96\linewidth]{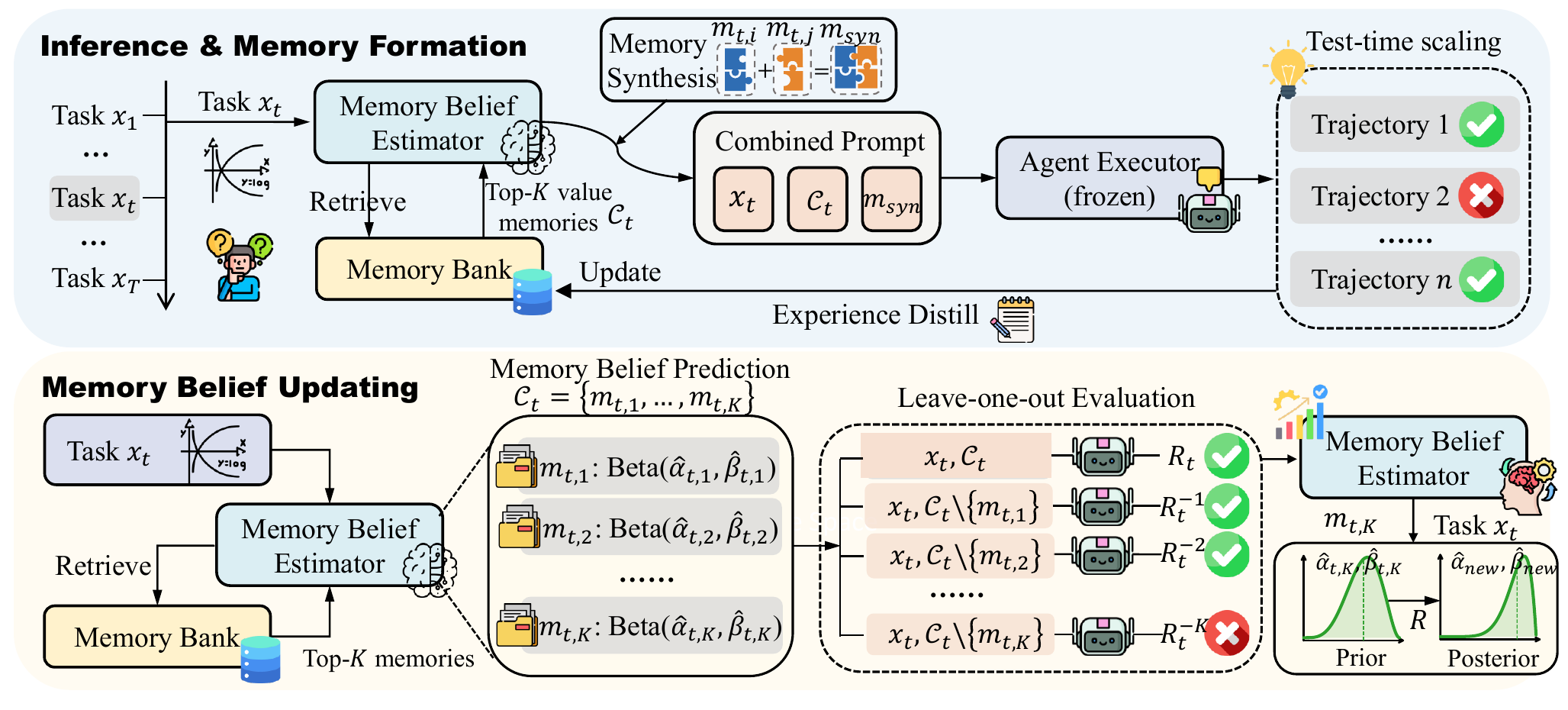}
  \caption{The overview of the BEAM.}
  \label{fig:overview}
\end{figure*}

\section{Methodology}
\subsection{Problem Setting}\label{sec:3.1}

Following~\citet{wei2025evo}, we consider a self-evolving LLM agent $\mathcal{A}_{\theta}$ that continuously solves a stream of tasks while maintaining an external reasoning memory. Let $\mathcal{D}=\{(x_t,y_t)\}_{t=1}^{T}$ denote the task sequence, where $x_t$ is the input context and $y_t$ is the corresponding target output. At step $t$, the agent retrieves a set of relevant memories $\mathcal{C}_t$ from the previous memory bank $M_{t-1}$ and generates a prediction:
\begin{equation}
\hat{y}_t=\mathcal{A}_{\theta}(x_t,\mathcal{C}_t).
\end{equation}
The objective of the self-evolving agent is to maximize its cumulative task utility over the task stream:
\begin{equation}
\mathcal{J}
=
\sum_{t=1}^{T}
R(\hat{y}_t,y_t)
=
\sum_{t=1}^{T}
R
\left(
\mathcal{A}_{\theta}(x_t,\mathcal{C}_t),y_t
\right)
,
\end{equation}
where $R(\cdot)$ denotes the task-specific utility of the prediction, such as answer correctness for reasoning tasks or execution success for coding tasks.

\subsection{Inference and Memory Formation}
We begin by detailing the forward task-solving process and the subsequent distillation of novel experiences, as illustrated in the upper pipeline of Figure~\ref{fig:overview}.

\paragraph{Memory-Augmented Inference.}
Given an incoming task $x_t$, the agent leverages a memory belief estimator $g_{\phi}(\cdot)$ (which will be formally defined in the next section) to infer the utility value of candidate memories in the external memory bank $\mathcal{M}$. Based on the estimated values, a set of top-$K$ memories $\mathcal{C}_t = \{m_{t,1}, \dots, m_{t,K}\}$ relevant to the current task context is retrieved. 
To better align these retrieved experiences with the specific requirements of $x_t$, the agent performs \textit{memory synthesis} to additionally extract a highly transferable working memory $m_{\mathrm{syn}}$ from $\mathcal{C}_t$. 
This synthesized working memory, together with the retrieved top-$K$ memories $\mathcal{C}_t$, is combined with $x_t$ to guide the frozen backbone LLM agent during reasoning, yielding the task execution response $\hat{y}_t$:
\begin{equation}
\hat{y}_t \sim \mathcal{A}_{\theta}(\cdot \mid x_t, \mathcal{C}_t, m_{\mathrm{syn}}) .
\end{equation}
Upon task completion, the environment returns a feedback signal $f_t$, producing a complete task interaction context.

\paragraph{Reasoning Memory Formation.}
To ensure high-quality experience abstraction and prevent redundant noise from contaminating long-term memory, we employ a \textit{test-time scaling} strategy during inference by sampling multiple execution trajectories $\{\tau_t^{(i)}\}_{i=1}^N$, where $\tau_t^{(i)} = (x_t, \hat{y}_t^{(i)}, f_t^{(i)})$. 
Conditioned on the environment feedback across these scaled trajectories, the agent reflects on the collective reasoning process, distilling actionable insights, such as transferable successful strategies or failure-aware lessons, into a structured reasoning memory entry:
\begin{equation}
m_t = (u_t, d_t, c_t),
\end{equation}
where $u_t$ denotes a concise memory title, $d_t$ summarizes the experience, and $c_t$ describes the reusable reasoning principle. The newly formed memory $m_t$, along with its context, is incorporated into the external memory bank $\mathcal{M}$ for future task retrieval and belief calibration.

\subsection{Context-conditioned Memory Belief Modeling}

Existing memory-augmented agents typically retrieve semantically relevant memories to facilitate task solving~\citep{rezazadeh2025isolated}, but fail to evaluate their reliability across changing task contexts. Recent studies estimate memory utility using scalar statistics, such as historical success rates~\citep{zhang2026memrl}. However, these approaches suffer from two main limitations. First, they assume memory value is task-independent, ignoring the fact that a memory's utility is highly context-dependent. For instance, a memory advocating \textit{"depth-first recursive traversal"} succeeds on small tasks but fails under strict stack constraints in large-scale settings. Second, scalar point estimates ignore evidence-based uncertainty, treating a single lucky success ($1/1$) identically to a proven record ($97/100$). Therefore, memory value should be modeled as a context-dependent belief that captures both contextual relevance and evidence-based uncertainty.

Inspired by the Bayesian brain hypothesis~\citep{knill2004bayesian,friston2010free}, which posits that intelligent systems continuously infer and update probabilistic beliefs by integrating prior knowledge with incoming evidence, we revisit memory value estimation from an evidential perspective. Instead of assigning each memory a deterministic utility score, we model its usefulness as a context-conditioned probabilistic belief. Specifically, given a task context $x_t$, we represent the latent utility of a candidate memory $m_{t,i}$ as a random variable:
\begin{equation}
v_{t,i}\sim \mathrm{Beta}(\alpha_{t,i},\beta_{t,i}),
\end{equation}
where $v_{t,i}$ denotes the latent probability that memory $m_{t,i}$ provides beneficial guidance for the current task $x_t$. We adopt the beta distribution, as it serves as the conjugate prior to the Bernoulli likelihood, which naturally aligns with the binary success-or-failure feedback of task execution. The parameters $\alpha_{t,i}$ and $\beta_{t,i}$ characterize the accumulated positive and negative evidence, respectively. This allows the agent to explicitly quantify both the expected memory utility and its associated uncertainty.

To parameterize the context-conditioned memory belief, we introduce a lightweight LoRA-adapted value estimator based on a pretrained language model. Rather than learning memory values directly, the estimator is trained to infer the parameters of the Beta belief distribution by jointly considering the current task context and candidate memory representations. Specifically, given the current task $x_t$ and a candidate memory $m_{t,i}$, the estimator produces the corresponding belief parameters:
\begin{equation}
\begin{aligned}
[\hat{\alpha}_{t,i},\hat{\beta}_{t,i}]
&=
\mathbf{1}
+
\operatorname{softplus}
\left(
g_{\phi}(x_t,m_i)
\right),
\end{aligned}
\label{eq:evidential_parameterization}
\end{equation}
where $g_{\phi}(\cdot)$ denotes the LoRA-adapted estimator with frozen backbone parameters, and $\hat{\alpha}_{t,i}$ and $\hat{\beta}_{t,i}$ represent the estimated parameters of the context-conditioned Beta belief. The softplus transformation ensures that the predicted evidence remains non-negative, while the unit offset corresponds to the uniform prior $\mathrm{Beta}(1,1)$ in the absence of supporting evidence. Consequently, when the estimator observes insufficient contextual evidence for a memory item, the resulting belief remains close to an uninformative distribution rather than producing an overconfident utility estimation. As more evidence is accumulated through interactions, the belief can gradually become more concentrated, reflecting increasing confidence in the reliability of the memory.

\subsection{Evidence-driven Memory Belief Updating}
The context-conditioned belief estimation above provides an initial prior over memory utility before the current interaction. However, such priors are inevitably uncertain due to incomplete contextual information and limited historical evidence. Similar to humans who continually revise beliefs through new experiences, we treat the feedback $f_t$ obtained after applying retrieved memories as empirical evidence to calibrate and refine prior beliefs, enabling memory values to continuously evolve with accumulated experience rather than remaining fixed after initial estimation.

Given a current task $x_t$ and a candidate memory $m_{t,i}$, the context-conditioned belief estimator models its usefulness as a belief distribution $\operatorname{Beta}(\hat{\alpha}_{t,i},\hat{\beta}_{t,i})$. To strike a balance between exploitation and exploration, we adopt Thompson sampling~\citep{thompson1933likelihood,chapelle2011empirical} to draw a stochastic utility score $\tilde{v}_{t,i}$ directly from this distribution:
\begin{equation}
\tilde{v}_{t,i}\sim
\operatorname{Beta}
(\hat{\alpha}_{t,i},\hat{\beta}_{t,i}).
\end{equation}

Unlike deterministic point estimates (e.g., predicted means), this sampling scheme intrinsically accounts for epistemic uncertainty. Consequently, memories with limited evidence retain higher variance, naturally yielding greater opportunities for exploration alongside the exploitation of well-supported experiences.

Since effective reasoning often benefits from complementary experiences~\citep{xu2026mem}, the agent retrieves a set of top-$K$ memories based on sampled utilities rather than selecting a single memory. 
However, the obtained feedback only reflects the joint effect of the retrieved memory set, making it difficult to determine the individual contribution of each memory. To address this challenge, we introduce leave-one-out counterfactual credit assignment to derive memory-level evidence.
For a task $x_t$, let $\mathcal{C}_t=\{m_{t,1},\ldots,m_{t,K}\}$ denote the retrieved memory set. We estimate the contribution of each memory by comparing the full retrieval outcome with leave-one-out counterfactual outcomes:
\begin{equation}
\begin{aligned}
R_t
&=
R(\mathcal{A}_{\theta}(x_t,\mathcal{C}_t),y_t),\\
R_t^{-i}
&=
R(\mathcal{A}_{\theta}(x_t,\mathcal{C}_t\setminus\{m_{t,i}\}),y_t),\\
A_{t,i}
&=
R_t-R_t^{-i},
\end{aligned}
\label{eq:counterfactual_credit}
\end{equation}
where $R(\cdot)\in\{0,1\}$ denotes the task reward ($1$ for success, $0$ otherwise), and $R_t^{-i}$ is the reward without memory $m_{t,i}$. The marginal contribution $A_{t,i}$ reflects $m_{t,i}$'s net impact, where positive and negative values indicate beneficial memories and harmful interference, respectively. To evaluate robustness across retrieval contexts, we aggregate outcomes retaining $m_{t,i}$ into an average counterfactual reward $\bar R_{t,i}^{\mathrm{ret}}=(K-1)^{-1}\sum_{j\neq i}R_t^{-j}$. 
We then combine the overall reward, counterfactual outcomes, and marginal contribution to construct the memory-level utility target:
\begin{equation}
y_{t,i}^{\mathrm{evi}}
=
\lambda \frac{
1+A_{t,i}
}{2}
+
(1-\lambda)
\frac{
R_t+\bar R_{t,i}^{\mathrm{ret}}
}{2},
\label{eq:credit_target}
\end{equation}
where $A_{t,i}\in\{-1,0,1\}$ is the marginal contribution of $m_{t,i}$. We normalize $A_{t,i}$ as $(1+A_{t,i})/2 \in \{1, 0.5, 0\}$ to align with the support of the Beta distribution, mapping positive, neutral, and negative impacts accordingly. The resulting target $y_{t,i}^{\mathrm{evi}}$ integrates both individual contribution and contextual effectiveness to supervise belief distribution calibration.

For the predicted distribution $\operatorname{Beta}(\hat{\alpha}_{t,i},\hat{\beta}_{t,i})$, let $\hat{\mu}_{t,i}$ and $\hat{\sigma}_{t,i}^{2}$ denote its mean and variance, respectively. We optimize the memory value estimator with an evidential objective:
\begin{equation}
\mathcal{L}_{\mathrm{evi}}
=
\frac{1}{|\mathcal{D}|}
\sum_{(t,i)\in\mathcal{D}}
\left[
(y_{t,i}^{\mathrm{evi}}-\hat{\mu}_{t,i})^2
+
\lambda_{\mathrm{var}}\hat{\sigma}_{t,i}^{2}
\right],
\label{eq:evidential_value_loss}
\end{equation}
where $\mathcal{D}$ denotes a training mini-batch. This objective aligns the predicted expected utility with evidence-derived supervision, while encouraging the predicted distribution to become more concentrated when supported by sufficient evidence.

However, calibrating absolute utility values alone does not guarantee preserving the relative preference among candidate memories retrieved for the same task. Since memory selection relies on ranking candidate experiences by their expected utilities, we further introduce a pairwise ranking objective based on counterfactual contributions. Specifically, for a mini-batch $\mathcal{D}$, we construct preference pairs $\mathcal{P}_{\mathcal D} = \{(t,i,j) \mid t \in \mathcal{D}, A_{t,i} > A_{t,j}\}$, where memory $m_{t,i}$ is preferred over $m_{t,j}$ due to its strictly higher marginal contribution. We optimize a margin-based ranking loss:
\begin{equation}
\mathcal{L}_{\mathrm{rank}}
=
\frac{1}{|\mathcal{P}_{\mathcal D}|}
\sum_{(t,i,j)\in\mathcal{P}_{\mathcal D}}
\left[
\rho-(\hat{\mu}_{t,i}-\hat{\mu}_{t,j})
\right]_{+},
\label{eq:counterfactual_ranking_loss}
\end{equation}
where $\rho>0$ is a margin enforcing separation between preferred pairs. This objective encourages memories with larger marginal contributions to yield higher utilities, preserving within-task preference ordering among experiences.

Due to noisy feedback in multi-memory interactions and limited observations, accumulated evidence can deviate from true memory utilities, leading to distorted belief strengths and poorly calibrated uncertainty estimates. To address this, we introduce a selective Beta-KL regularizer that suppresses unsupported evidence accumulation while preserving evidence consistent with the utility target. Specifically, we isolate potentially conflicting evidence for selective regularization:
\begin{equation}
\widetilde{\alpha}_{t,i}
=
1+
(1-y_{t,i}^{\mathrm{evi}})
(\hat{\alpha}_{t,i}-1),
\qquad
\widetilde{\beta}_{t,i}
=
1+
y_{t,i}^{\mathrm{evi}}
(\hat{\beta}_{t,i}-1),
\end{equation}
where $\hat{\alpha}_{t,i}-1$ and $\hat{\beta}_{t,i}-1$ denote the accumulated positive and negative evidence underlying the predicted Beta distribution, respectively. When $y_{t,i}^{\mathrm{evi}}$ is high, negative evidence is regarded as conflicting and selectively regularized; conversely, when $y_{t,i}^{\mathrm{evi}}$ is low, excessive positive evidence is suppressed. For intermediate targets, both evidence terms are softly weighted according to the utility target. The selective regularization objective is formulated as:
\begin{equation}
\mathcal{L}_{\mathrm{kl}}
=
\frac{1}{|\mathcal{D}|}
\sum_{(t,i)\in\mathcal{D}}
D_{\mathrm{KL}}
\left[
\operatorname{Beta}
(\widetilde{\alpha}_{t,i},\widetilde{\beta}_{t,i})
\,\middle\|\,
\operatorname{Beta}(1,1)
\right],
\label{eq:selective_beta_kl}
\end{equation}
where $\operatorname{Beta}(1,1)$ denotes the uninformative prior representing zero accumulated evidence. This objective selectively contracts target-inconsistent evidence toward the prior, preventing unsupported belief accumulation and improving the calibration of memory value estimates.

The overall training objective jointly optimizes evidential utility calibration, preference-aware ranking, and selective evidence regularization:
\begin{equation}
\mathcal{L}_{\mathrm{value}}
=
\mathcal{L}_{\mathrm{evi}}
+
\lambda_{\mathrm{rank}}\mathcal{L}_{\mathrm{rank}}
+
\lambda_{\mathrm{kl}}\mathcal{L}_{\mathrm{kl}},
\label{eq:overall_value_objective}
\end{equation}
where $\lambda_{\mathrm{rank}}$ and $\lambda_{\mathrm{kl}}$ are trade-off hyperparameters controlling the contributions of the ranking and evidence regularization objectives, respectively. We further assess the contribution of each loss term through individual ablations, with the results reported in the supplementary material.

\section{Experiments} \label{sec:sec5}

\begin{table*}[t]
    \centering
    \setlength{\tabcolsep}{5.2pt}

    \begin{tabular*}{\textwidth}{@{\extracolsep{\fill}}lcccccc@{}}
        \specialrule{1.4pt}{0pt}{0.5pt}

        \multicolumn{1}{c}{}
        & \multicolumn{4}{c}{\textbf{Accuracy (\%) $\uparrow$}}
        & \multicolumn{1}{c}{\textbf{Pass@1 (\%) $\uparrow$}}
        & \multicolumn{1}{c}{} \\
        \cmidrule(lr){2-5}
        \cmidrule(lr){6-6}
        \textbf{Method}
        & \textbf{AIME}
        & \textbf{GPQA-Diamond}
        & \textbf{HLE}
        & \textbf{MMLU-Pro (Philo.)}
        & \textbf{BigCodeBench}
        & \textbf{Avg. $\uparrow$} \\
        \midrule

        \rowcolor{black!8}
        \multicolumn{7}{c}{\rule[-0.45ex]{0pt}{2.6ex}\textbf{DeepSeek-V4-Flash}} \\
        No Memory & 58.33 & 69.10 & 9.09 & 73.61 & 31.08 & 48.24 \\
        Self-RAG  & 56.67 & 70.70 & 10.83 & 71.83 & 31.08 & 48.22 \\
        Mem0      & 61.67 & 69.70 & 11.61 & 72.62 & 29.73 & 49.07 \\
        A-Mem     & 56.67 & 70.71 & 11.99 & \textbf{77.78} & 30.41 & 49.51 \\
        Memp      & 60.00 & 68.18 & 7.93 & 75.79 & 28.38 & 48.06 \\
        MemRL     & 58.33 & 69.70 & 11.03 & 75.79 & 28.38 & 48.65 \\
        BEAM (Ours)      & \textbf{66.67} & \textbf{75.25} & \textbf{12.77} & 77.38 & \textbf{35.13} & \textbf{53.44} \\

        \specialrule{0.55pt}{1.2pt}{1.2pt}

        \rowcolor{black!8}
        \multicolumn{7}{c}{\rule[-0.45ex]{0pt}{2.6ex}\textbf{Gemini-2.5-Flash}} \\
        No Memory & 53.33 & 73.23 & 12.96 & 77.58 & 25.68 & 48.56 \\
        Self-RAG  & 56.67 & 71.72 & 13.35 & 78.37 & 24.32 & 48.89 \\
        Mem0      & 58.33 & 76.76 & 11.99 & 78.77 & 25.68 & 50.31 \\
        A-Mem     & 63.33 & 77.78 & 15.09 & 76.79 & 27.70 & 52.14 \\
        Memp      & 53.33 & 74.75 & 10.83 & 78.37 & 26.35 & 48.73 \\
        MemRL     & 56.67 & 75.25 & 10.44 & 77.78 & 27.03 & 49.43 \\
        BEAM (Ours)      & \textbf{68.33} & \textbf{78.79} & \textbf{15.47} & \textbf{79.17} & \textbf{31.08} & \textbf{54.57} \\

        \specialrule{1.4pt}{0.5pt}{0pt}
    \end{tabular*}
    \caption{Performance comparison of memory methods across five benchmarks and two fixed task executors. Scores are accuracy except on BigCodeBench, where Pass@1 is reported. Avg. denotes the arithmetic mean across the five benchmarks.}
    \label{tab:main_results}
\end{table*}

\subsection{Experiment Settings}

\textbf{Datasets.} We evaluate BEAM on five benchmarks spanning mathematical reasoning, using AIME24 and AIME25~\citep{aime24,aime25}; scientific reasoning, using GPQA-Diamond~\citep{rein2023gpqa}; philosophical reasoning, using MMLU-Pro (Philosophy)~\citep{wang2024mmlu}; multidisciplinary reasoning, using the valid text-only multiple-choice subset of HLE~\citep{phan2025humanity}; and code generation, using BigCodeBench-Hard-Instruct~\citep{zhuo2025bigcodebench}. We report accuracy for reasoning benchmarks and Pass@1 for BigCodeBench.

\noindent\textbf{Baselines.} We compare BEAM with six representative baselines spanning diverse memory paradigms: No Memory, which solves tasks using only the frozen backbone language model; Self-RAG~\citep{asai2024self}, which controls retrieval through self-reflection; Mem0~\citep{chhikara2025mem0}, a general-purpose long-term memory framework; A-Mem~\citep{xu2026mem}, which organizes structured memory notes and semantic associations; Memp~\citep{fang2026memp}, which distills procedural knowledge from trajectories; and MemRL~\citep{zhang2026memrl}, which combines semantic retrieval with learned Q-values. All methods use the same task streams, executors, output parsers, and evaluation protocols while preserving each baseline's core memory mechanisms.

\noindent\textbf{Evaluation Protocol.} Following Evo-Memory's streaming search--predict--evolve protocol~\citep{wei2025evo}, we adopt a zero-reset task-stream setting. For each benchmark, the corresponding task stream is partitioned into consecutive stages and processed in a strictly fixed order, without resetting the memory bank or value estimator at any point throughout the stream. For each task, inference relies exclusively on experience accumulated from preceding tasks, while feedback from the current task is used only to update the system for subsequent tasks, preventing it from influencing the current prediction. Additional implementation details, including the key prompts used throughout BEAM, are provided in the supplementary material.

\subsection{Main Results and Analysis}

Table~\ref{tab:main_results} compares memory methods across two executors and five benchmarks. BEAM achieves the best average performance under both executors, surpassing the strongest baselines by 3.93 and 2.43 percentage points, respectively. It outperforms all baselines on AIME, GPQA-Diamond, HLE, and BigCodeBench, while remaining competitive on MMLU-Pro. These consistent gains across reasoning and code-generation tasks show that BEAM generalizes across the two tested executors and diverse task settings, rather than relying on a specific backbone or evaluation scenario. They further demonstrate its effectiveness on knowledge-intensive problems, where the utility of past experience varies substantially across instances.

More broadly, the baseline comparison shows that external memory does not necessarily yield consistent gains. Several methods vary substantially across tasks, and some even underperform No Memory, suggesting that semantic relevance alone may retrieve seemingly related memories with limited practical utility for current tasks, increasing the risk of negative transfer. Although A-Mem and MemRL improve memory use through structured organization and value-aware selection, respectively, their overall performance remains below that of BEAM. In contrast, BEAM models the task-conditioned utility of past experience through probabilistic value estimation and continually calibrates candidate-memory rankings using counterfactual feedback, yielding more consistent gains across both executors and diverse tasks. Two qualitative case studies illustrating BEAM's end-to-end reasoning and memory-utilization process are provided in the supplementary material.

\subsection{Ablation Studies}

\begin{figure}[t]
    \centering
    \includegraphics[width=\columnwidth]{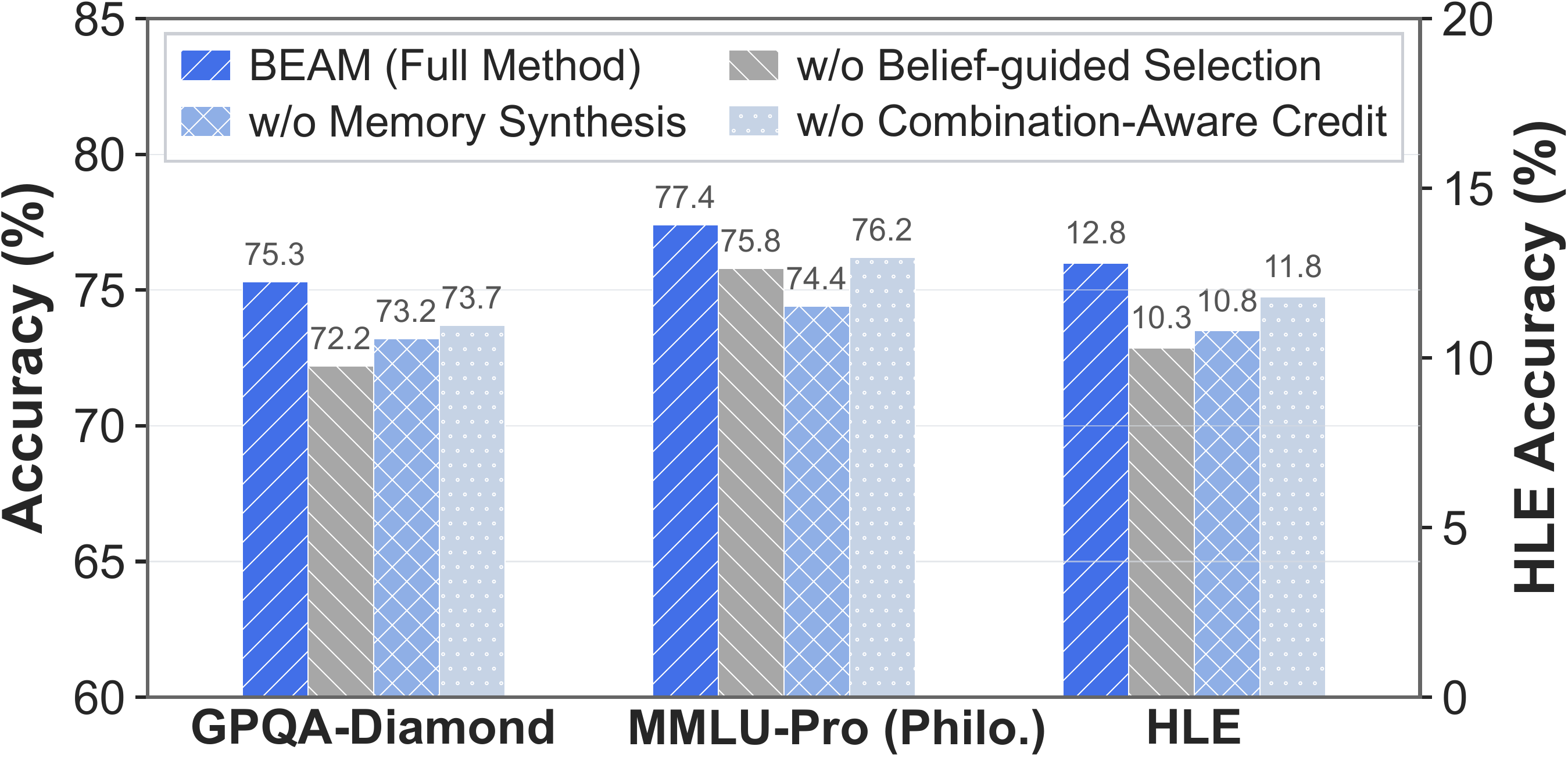}
    \caption{Ablation results on GPQA-Diamond, MMLU-Pro (Philo.), and HLE. The left axis reports accuracy on GPQA-Diamond and MMLU-Pro, while the right axis reports accuracy on HLE.}
    \label{fig:component_ablation}
\end{figure}

As shown in Figure~\ref{fig:component_ablation}, we assess BEAM's key components on DeepSeek-V4-Flash across GPQA-Diamond, MMLU-Pro (Philo.), and HLE. To further assess the impact of test-time scaling, we compare different memory methods under the same three-rollout inference budget, using either random selection (Random-of-3) or a shared judge to select the best output (Best-of-3); the corresponding results are reported in the supplementary material.

\paragraph{Belief-guided Memory Selection.}
When belief-guided selection is removed, the system selects memories solely according to semantic similarity, resulting in consistent performance degradation across all three benchmarks. This indicates that semantic relevance alone is insufficient to characterize the actual utility of a memory for the current task. In contrast, task-conditioned belief estimation helps filter out memories that are superficially relevant but only marginally useful or potentially misleading.

\paragraph{Memory Synthesis.}
When memory synthesis is removed, the system still injects the selected raw memories but no longer constructs an additional working memory. The resulting performance drop suggests that raw memories may still contain redundancy, conflicts, and task-irrelevant details. By distilling transferable reasoning principles and integrating complementary information, memory synthesis provides the executor with more focused and task-conditioned guidance.

\paragraph{Combination-Aware Credit Assignment.}
BEAM uses leave-one-out evaluation to estimate the conditional marginal contribution of each memory within the actual multi-memory context, whereas the ablated variant compares each individual memory against an empty context:
\begin{equation}
A_{t,i}
=
R_t^{\mathrm{full}}-R_t^{-i},
\qquad
A_{t,i}^{\mathrm{single}}
=
R_t^{i}-R_t^{\mathrm{empty}}.
\end{equation}
The latter measures only the effect of a memory when used in isolation, while ignoring interactions among memories in the actual multi-memory reasoning context. As a result, it performs worse on all three benchmarks. These results show that, compared with isolated evaluation, combination-aware counterfactual supervision provides more reliable training signals for the value model.

\subsection{Continual Self-Evolution}
To assess whether BEAM can progressively improve memory use through continual interaction, we partition each task stream into consecutive stages and report stage-wise accuracy and the gain over the no-memory baseline:
$\Delta=\mathrm{Acc}_{\mathrm{method}}-\mathrm{Acc}_{\mathrm{no\text{-}memory}}$.
Neither the memory bank nor the value model is reset throughout the stream. Leave-one-out supervision from completed tasks is used only to update the value model for subsequent tasks, preventing it from affecting the current prediction.

\begin{figure}[t]
    \centering
    \includegraphics[width=\columnwidth]
    {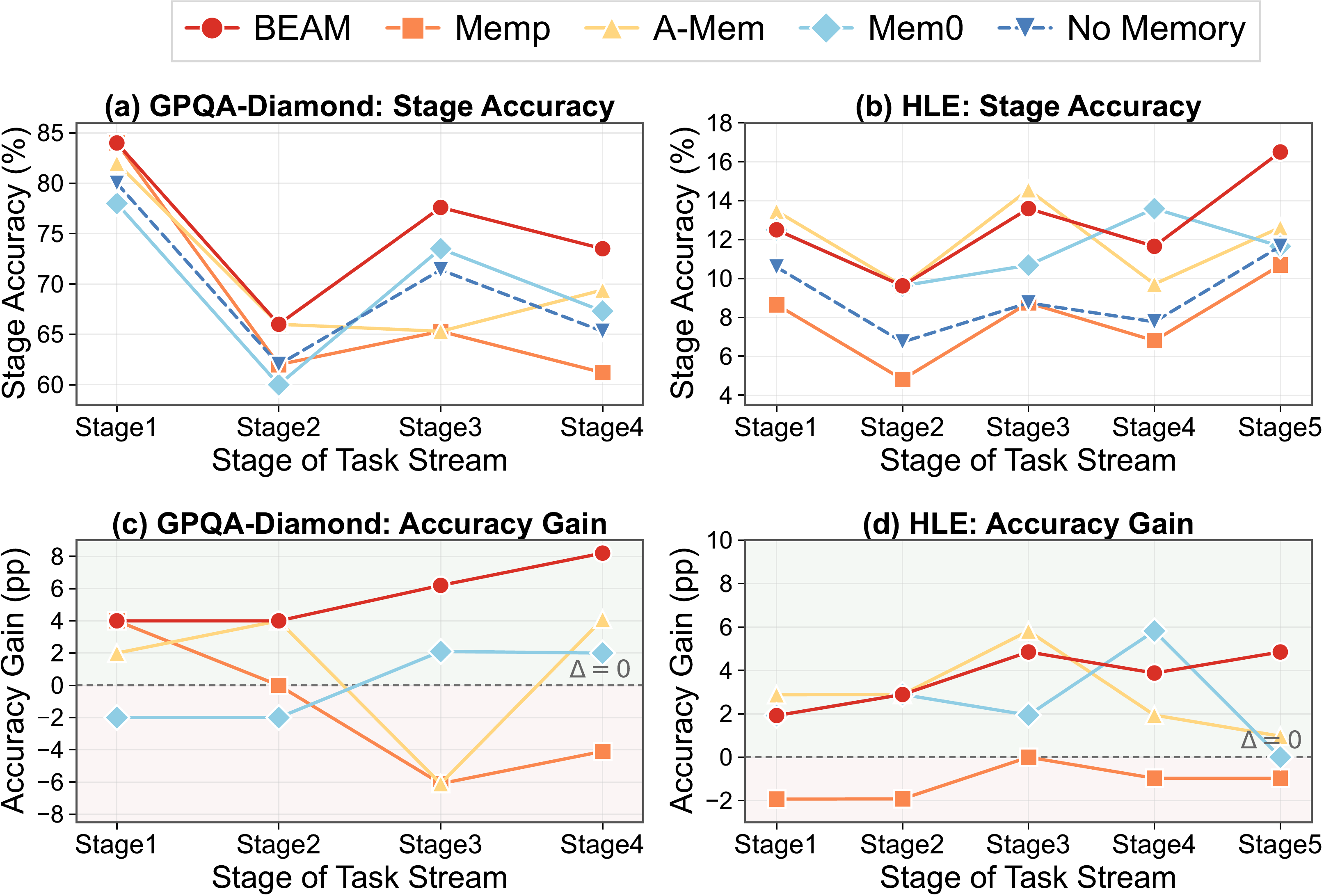}
    \caption{Stage-wise self-evolution on GPQA-Diamond and HLE. Each stage is a consecutive segment of the task stream; GPQA-Diamond uses four stages (0--25\%, 25--50\%, 50--75\%, 75--100\%), and HLE uses five. Panels (a,b) report stage accuracy, while panels (c,d) show gains over no memory in percentage points (pp), $\Delta=\mathrm{Acc}_{\mathrm{method}}-\mathrm{Acc}_{\mathrm{no\text{-}memory}}$.}
    \label{fig:stage_evolution}
\end{figure}

As shown in Figure~\ref{fig:stage_evolution}, the stage-wise accuracy of all methods fluctuates because task difficulty varies across stages. Nevertheless, BEAM generally achieves larger gains over the no-memory baseline as the task stream progresses, with more pronounced improvements in later stages. In contrast, other memory-based methods exhibit less stable gains, with some falling below the no-memory baseline at certain stages. This suggests that continual experience accumulation alone does not guarantee sustained improvement; memory quality and selection remain critical. As counterfactual supervision accumulates, BEAM's belief model progressively calibrates candidate-memory utility estimates, increasing the likelihood of selecting task-relevant memories while reducing interference from mismatched ones. Although this trend may be affected by stage difficulty and finite-sample variability, the larger relative gains in later stages provide empirical evidence that BEAM gradually improves memory utilization through continual self-evolution.

\subsection{Cross-Task Transferability}

\begin{figure}[t]
    \centering
    \includegraphics[width=\columnwidth]{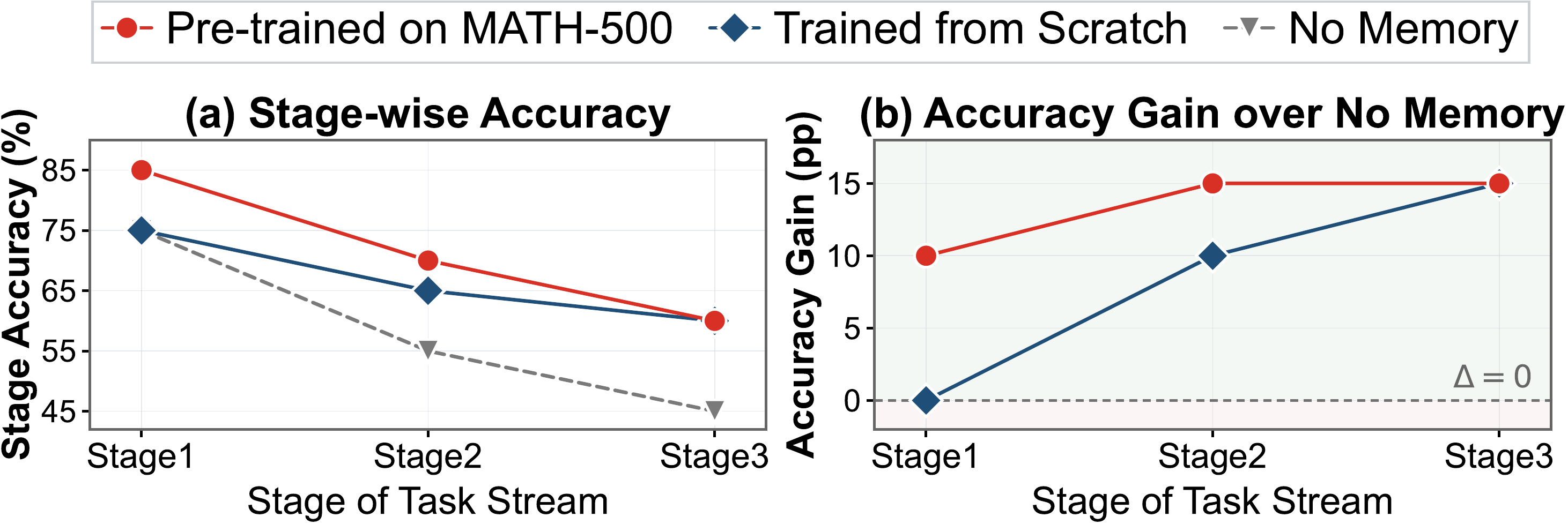}
    \caption{Cross-task transfer from MATH-500 to AIME. The AIME task stream is divided into three consecutive stages (0--33\%, 33--67\%, 67--100\%). Panel (a) reports stage accuracy, while panel (b) reports gains over no memory in percentage points (pp), $\Delta=\mathrm{Acc}_{\mathrm{method}}-\mathrm{Acc}_{\mathrm{no\text{-}memory}}$.}
    \label{fig:transfer_learning}
\end{figure}

To examine whether BEAM's value model transfers across related tasks, we initialize the AIME value network with parameters learned on MATH-500~\citep{hendrycks2021measuring,lightman2024let} and compare it with a value model trained from scratch on AIME and a no-memory baseline. As shown in Figure~\ref{fig:transfer_learning}, the transferred model achieves the best stream-level average performance and consistently outperforms the no-memory baseline across all stages. In particular, it shows a clear early-stage advantage over training from scratch, indicating that memory-utility judgments learned on MATH-500 provide an effective initialization for selecting useful memories on AIME when target-task feedback is still scarce. As AIME feedback accumulates, the scratch-trained model narrows the gap and achieves a comparable gain in the final stage. These results suggest that cross-task transfer improves early adaptation, feedback efficiency, and overall stream-level performance, while having only a limited effect on final-stage performance.

\section{Conclusion}

In this paper, we introduced BEAM, a framework that reformulates memory valuation for self-evolving LLM agents as inference over context-conditioned probabilistic beliefs. Rather than assigning each memory a static scalar utility, BEAM jointly models its expected usefulness and uncertainty, and continually calibrates these beliefs using execution feedback and counterfactual credit assignment. By integrating structured reasoning memory formation, belief-guided selection, and evidence-driven updating while keeping the backbone LLM frozen, BEAM enables reliable experience reuse across continual task streams. Experiments on diverse reasoning and code-generation benchmarks show that BEAM achieves consistent gains over strong memory baselines, progressively improves memory utilization through online interaction, and transfers learned utility judgments across related tasks. These findings underscore context-aware memory valuation as a promising foundation for reliable and continual self-evolution in LLM agents.

\bibliography{aaai2027}

\newpage
\begin{center}
    \huge \textbf{Appendix} \label{sec:appendix}
\end{center}
\appendix

\section{Implementation Details}

The memory belief estimator is initialized from Qwen2.5-7B-Instruct and equipped with a lightweight evidential Beta head that predicts context-conditioned belief parameters $(\alpha,\beta)$ for each task--memory pair. The pretrained backbone remains frozen, while LoRA adapters are applied to the query, key, value, and output projection layers with rank $r=8$, scaling factor $\alpha_{\mathrm{LoRA}}=16$, and dropout $0.05$. Pairwise inputs are encoded with a maximum sequence length of 4,096 tokens. Dense retrieval is performed using \texttt{text-embedding-3-large} embeddings stored in a local Qdrant vector database. For each task, six candidate memories are retrieved, and semantic similarity is combined with Thompson-sampled memory utility after z-score normalization. The value weight $\eta$ follows a sigmoid schedule over $[0.10,0.50]$ according to task-stream progress. After reranking, up to three memories with semantic similarity scores above the injection threshold of $0.5$ are retained for subsequent reasoning. The estimator is optimized online using AdamW, with learning rates of $5\times10^{-5}$ for the LoRA adapters and $1\times10^{-3}$ for the evidential Beta head, and a batch size of 8. The weights of $\mathcal{L}_{\mathrm{evi}}$, $\mathcal{L}_{\mathrm{kl}}$, and $\mathcal{L}_{\mathrm{rank}}$ are set to $1.0$, $0.03$, and $0.6$, respectively. Experiments are conducted on a Linux server equipped with multiple NVIDIA GeForce RTX 4090 GPUs. The memory belief estimator runs locally on one of the GPUs using bfloat16 precision, while external LLM and embedding services are accessed via APIs.

\section{Contribution of Value-Model Training Objectives}

\begin{figure}[!htbp]
    \centering
    \includegraphics[width=\columnwidth]
    {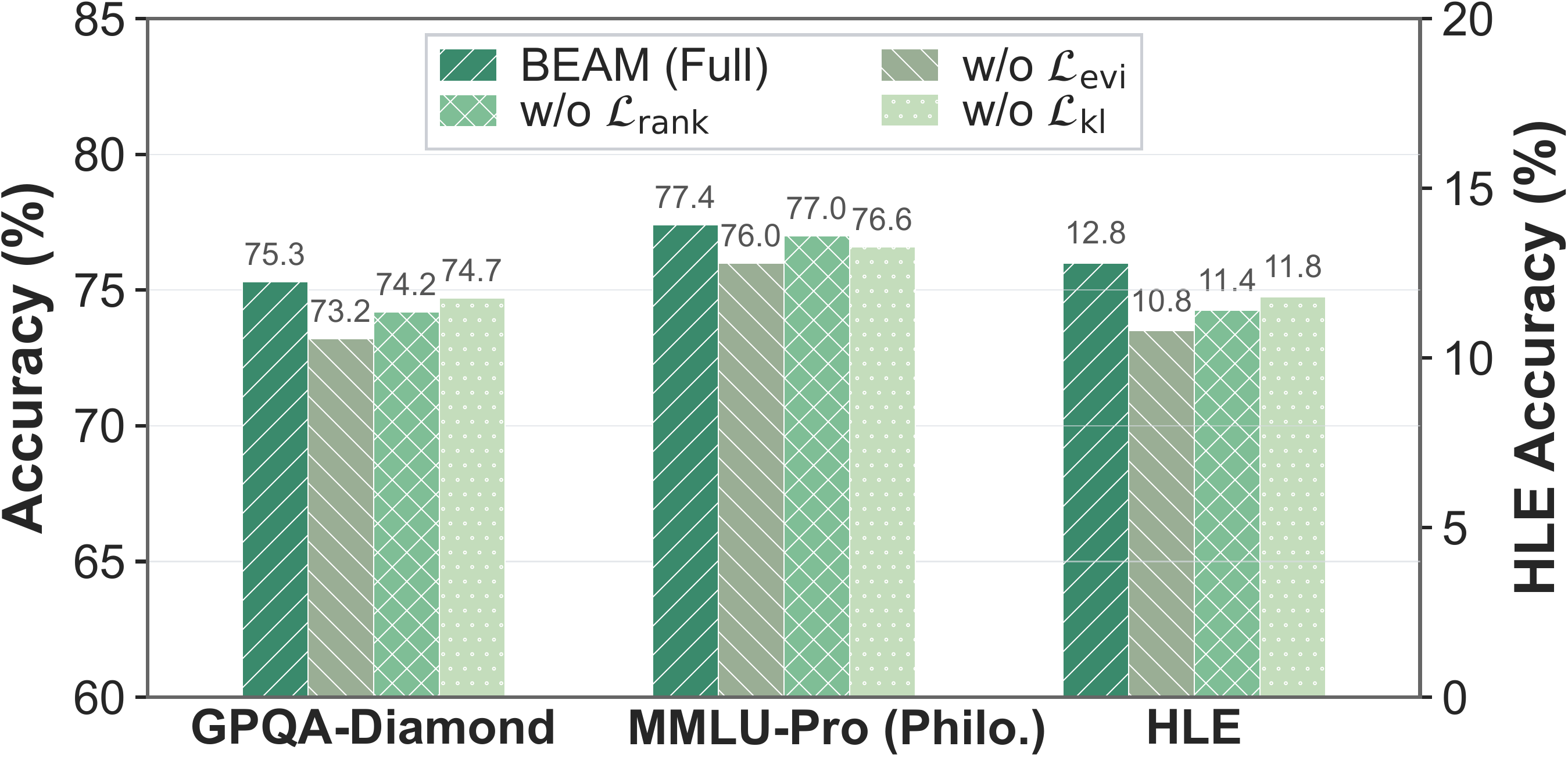}
    \caption{Ablation of the value-model training objectives on
    GPQA-Diamond, MMLU-Pro (Philo.), and HLE. The left y-axis reports
    accuracy on GPQA-Diamond and MMLU-Pro (Philo.), while the
    right y-axis reports accuracy on HLE.}
    \label{fig:loss_ablation}
\end{figure}

As shown in Figure~\ref{fig:loss_ablation}, we evaluate the contribution of each value-model objective by individually removing $\mathcal{L}_{\mathrm{evi}}$, $\mathcal{L}_{\mathrm{rank}}$, or $\mathcal{L}_{\mathrm{kl}}$. The full objective achieves the best performance on GPQA-Diamond, MMLU-Pro (Philo.), and HLE, while removing any individual loss consistently degrades performance, confirming their complementary roles.

Removing $\mathcal{L}_{\mathrm{evi}}$ causes the largest decline, with an average drop of 1.83 percentage points across the three benchmarks. This highlights the importance of evidential supervision for calibrating predicted mean utility and establishing a stable task-conditioned value scale. Without this objective, the model can still learn relative preferences through ranking supervision but lacks stable absolute utility estimates on a common scale.

Removing $\mathcal{L}_{\mathrm{rank}}$ also consistently reduces performance, indicating that fitting individual utility targets alone does not guarantee correct candidate ordering within the same task. Explicit preference constraints improve the separation between high- and low-utility memories, thereby supporting more reliable final selection. Removing $\mathcal{L}_{\mathrm{kl}}$ yields a smaller but consistent decline, supporting its role in regularizing unsupported evidence, mitigating overconfidence, and stabilizing belief estimation. Overall, $\mathcal{L}_{\mathrm{evi}}$ provides absolute utility calibration, $\mathcal{L}_{\mathrm{rank}}$ strengthens candidate discrimination, and $\mathcal{L}_{\mathrm{kl}}$ regularizes uncertain beliefs, jointly enabling more accurate and stable memory utility estimation.

\section{Effect of Test-Time Scaling}

\begin{figure}[!htbp]
    \centering
    \includegraphics[width=\columnwidth]
    {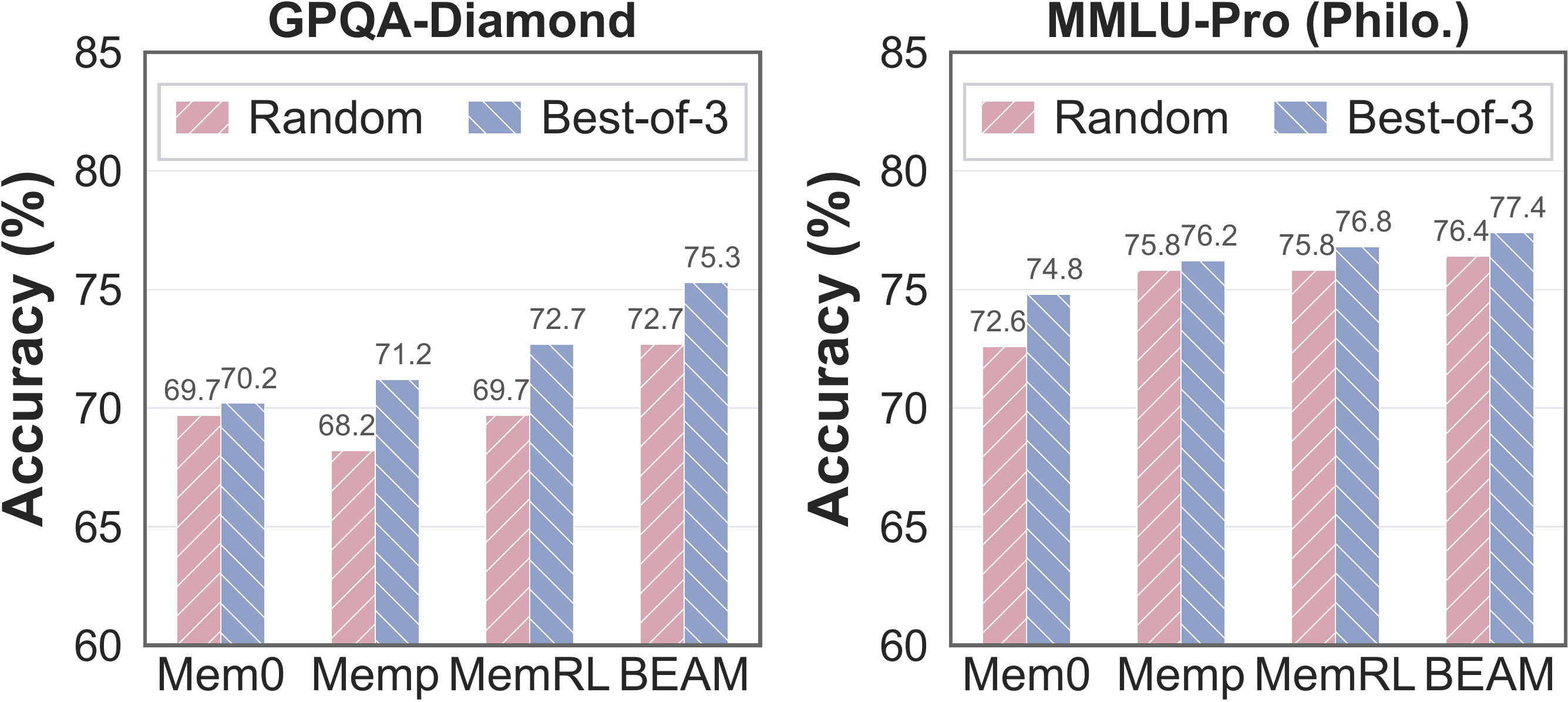}
    \caption{Comparison of Random-of-3 and Best-of-3 inference across memory methods on GPQA-Diamond and MMLU-Pro (Philo.).}
    \label{fig:selection_comparison}
\end{figure}

BEAM employs memory-augmented test-time scaling (MATTS), generating multiple memory-conditioned reasoning trajectories and using a shared judge to select the output. This raises the possibility that part of the gain comes from additional test-time computation rather than memory selection itself. To control for this factor, we allocate the same three-rollout budget to Mem0, Memp, and MemRL, and report Random-of-3 and Best-of-3, using random selection and the same judge, respectively. As shown in Figure~\ref{fig:selection_comparison}, Best-of-3 outperforms Random-of-3 in most settings, confirming that test-time scaling provides additional benefits. Nevertheless, under the same inference budget and selection mechanism, all baselines remain below BEAM on GPQA-Diamond and MMLU-Pro (Philo.). Moreover, BEAM's Random-of-3 performance already matches or approaches that of the strongest Best-of-3 baseline. These results indicate that repeated sampling is not the primary source of BEAM's performance advantage: high-quality memory selection and synthesis first improve the quality of candidate trajectories, while MATTS further mitigates the randomness and instability of rollouts.

\begin{figure*}[!t]
    \centering
    \includegraphics[width=\textwidth]
    {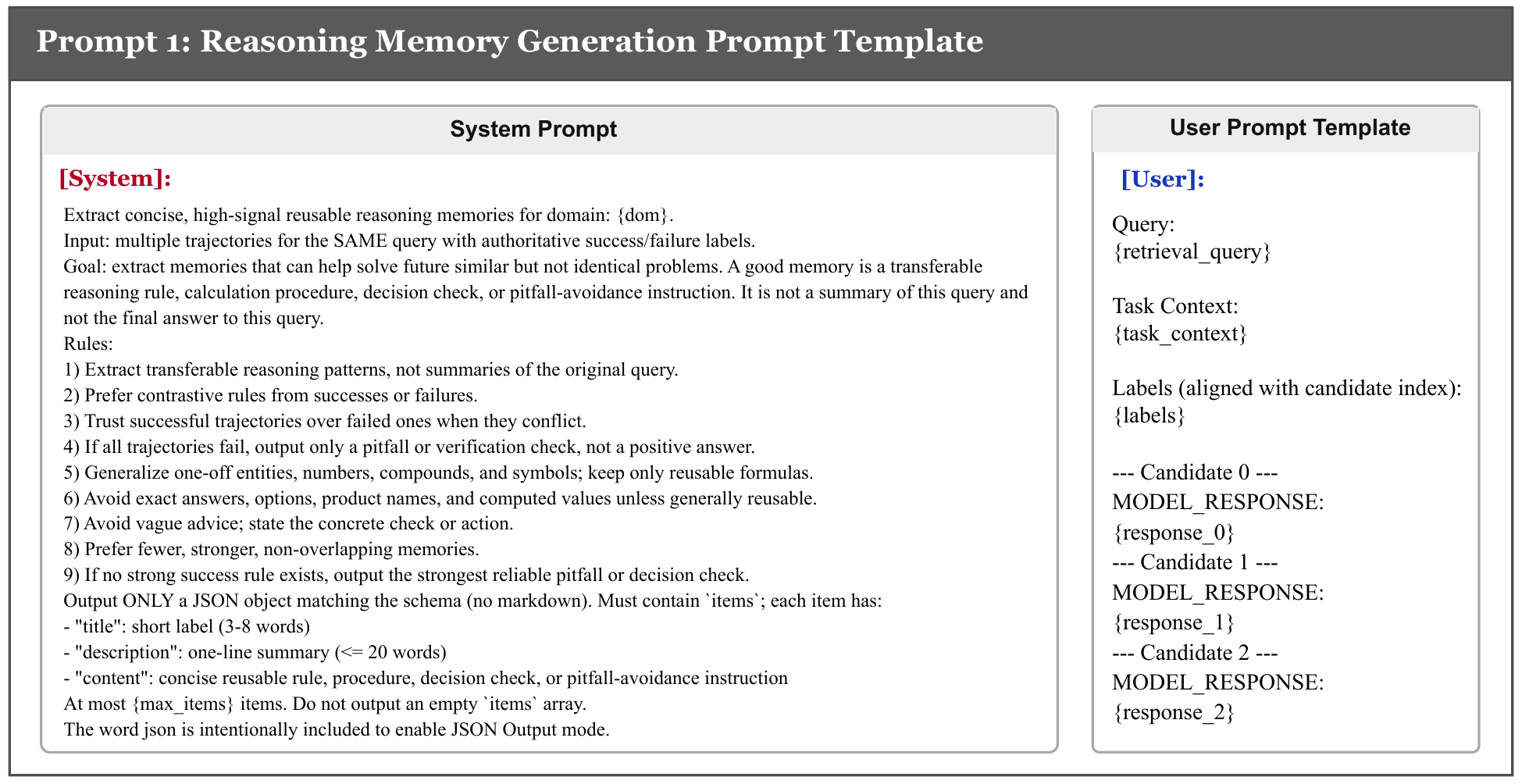}
    \caption{System and user prompt templates for reasoning memory generation.}
    \label{fig:prompt_memory_generation}
\end{figure*}

\begin{figure*}[!t]
    \centering
    \includegraphics[width=\textwidth]
    {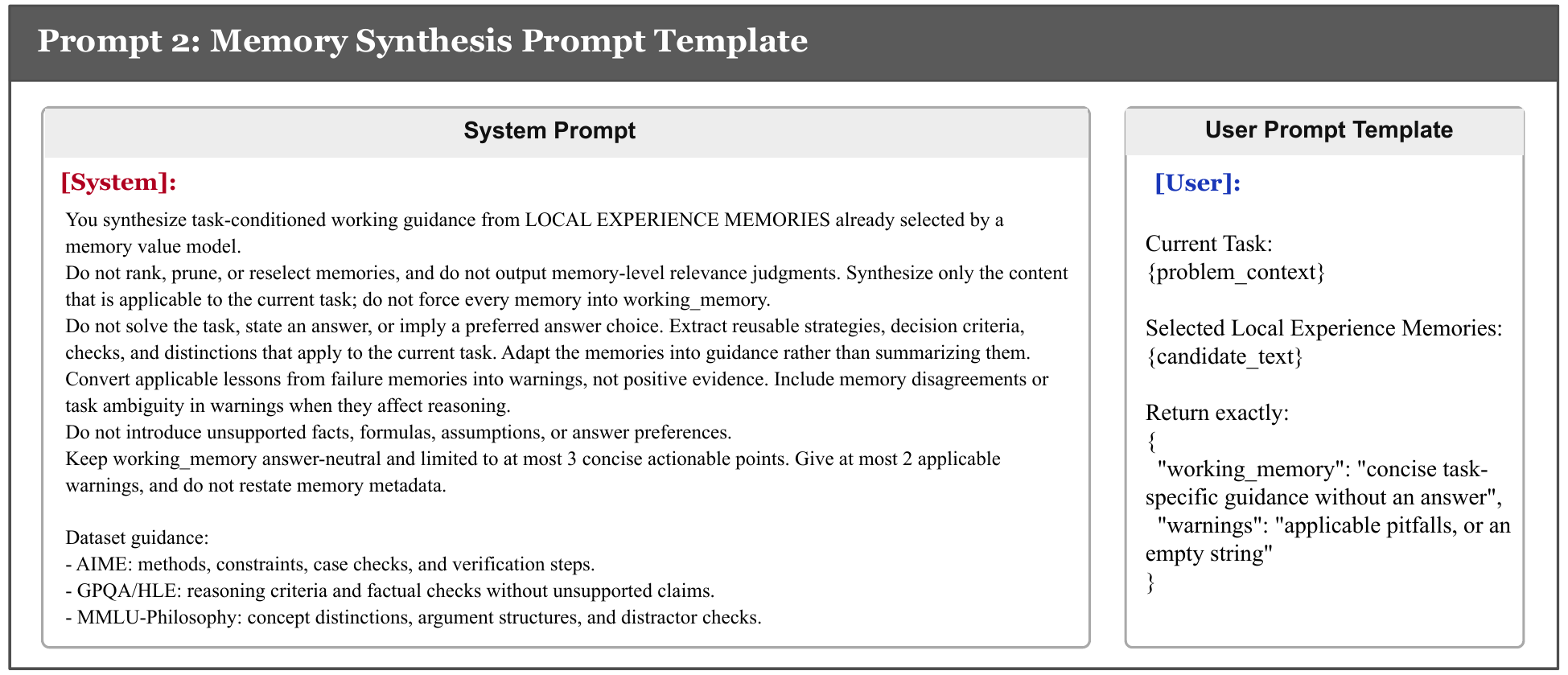}
    \caption{System and user prompt templates for memory synthesis.}
    \label{fig:prompt_memory_synthesis}
\end{figure*}

\section{Prompt Templates}
We present the key prompts used for reasoning memory generation and memory synthesis in BEAM. Both use structured JSON outputs to support reliable downstream processing.

\paragraph{Reasoning Memory Generation.}
Figure~\ref{fig:prompt_memory_generation} shows how BEAM extracts reusable memories from multiple labeled trajectories of the same task. The prompt prioritizes patterns supported by successful trajectories and converts failures into pitfalls or verification checks; when all trajectories fail, only cautionary guidance is produced. Instance-specific details are generalized, and each memory is returned with structured \texttt{title}, \texttt{description}, and \texttt{content} fields.

\paragraph{Memory Synthesis.}
Figure~\ref{fig:prompt_memory_synthesis} presents the prompt that generates an answer-neutral, task-conditioned working memory from the current task and the memories selected by the value model, without reranking the memories or solving the task. Applicable lessons from failed memories and conflicts among memories are expressed as warnings. The output is organized into \texttt{working\_memory} and \texttt{warnings}.

\begin{figure*}[!t]
    \centering
    \includegraphics[width=\textwidth]
    {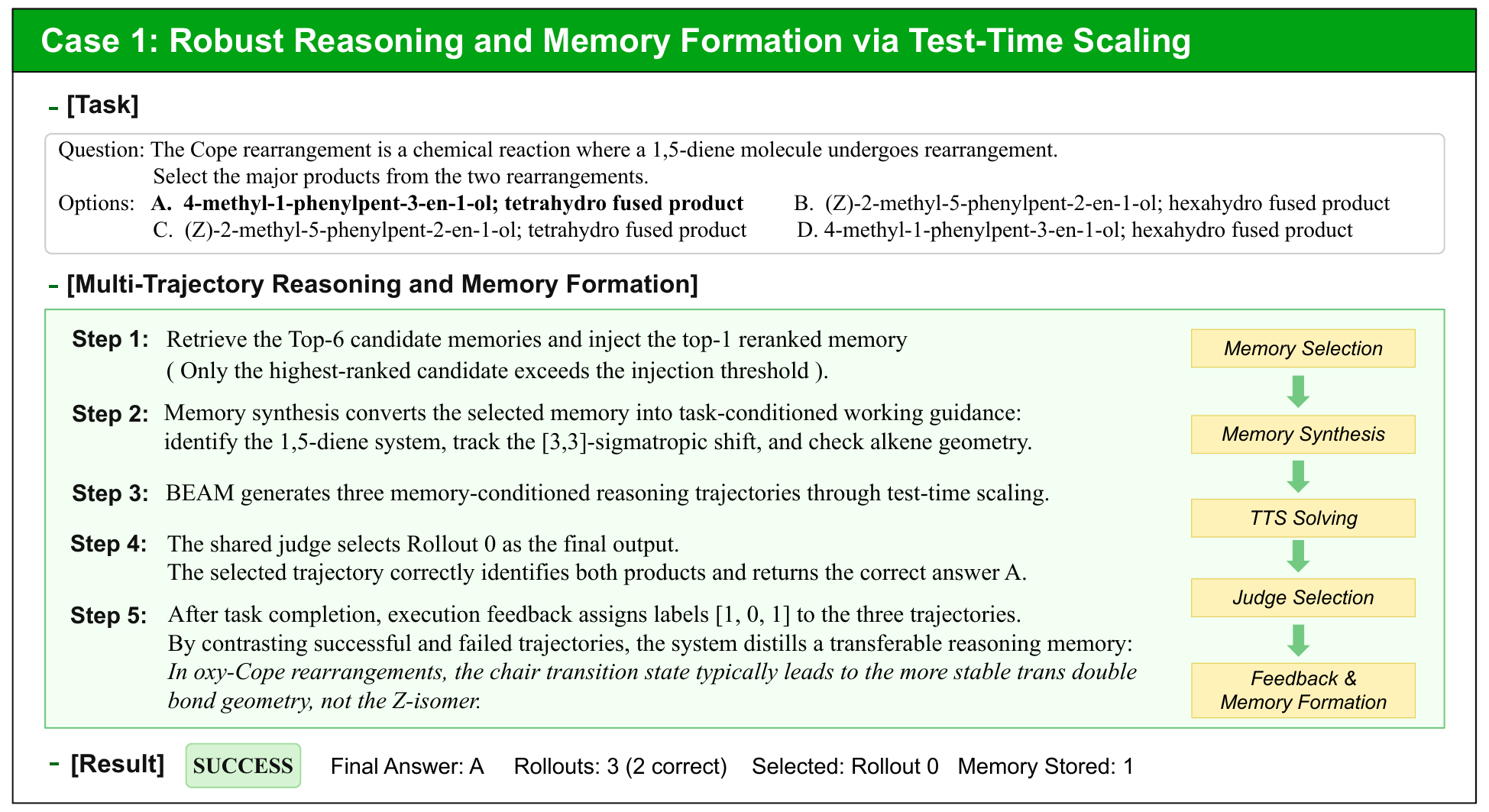}
    \caption{Case study illustrating BEAM's test-time scaling and feedback-driven memory formation. The task text and answer choices are abridged for readability.}
    \label{fig:case_tts}
\end{figure*}

\begin{figure*}[!t]
    \centering
    \includegraphics[width=\textwidth]
    {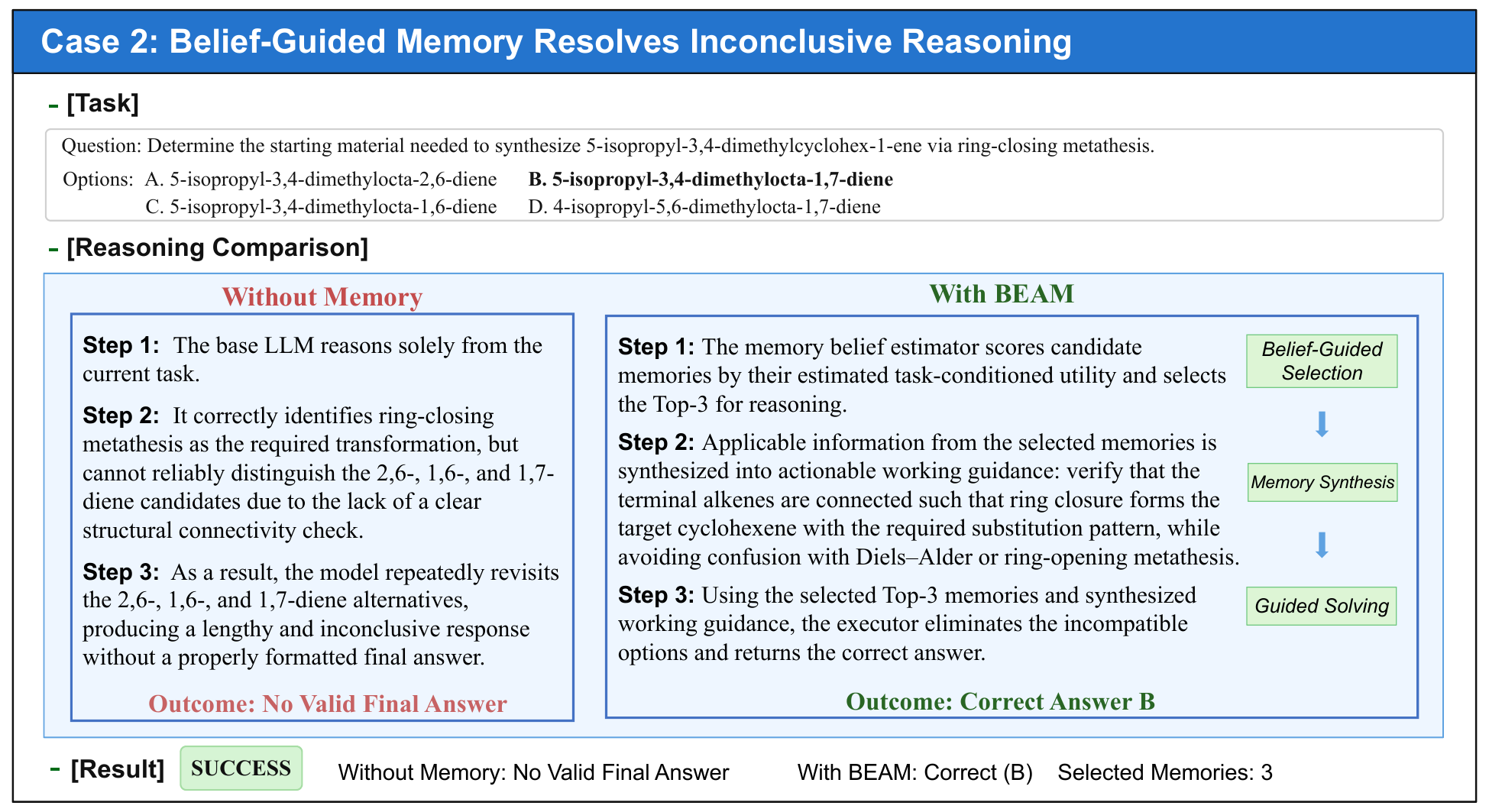}
    \caption{Case study illustrating BEAM's belief-guided memory selection and task-conditioned memory synthesis. The task text and answer choices are abridged for readability.}
    \label{fig:case_memory_utilization}
\end{figure*}

\section{Case Studies}

We present two cases illustrating BEAM's closed-loop self-evolution process. The first shows how robust multi-trajectory reasoning supports reliable memory formation, while the second demonstrates how task-conditioned memory utilization subsequently improves reasoning accuracy.

\paragraph{Case 1: Robust Reasoning and Memory Formation.}
As shown in Figure~\ref{fig:case_tts}, BEAM generates multiple memory-conditioned trajectories and uses a shared judge to select a reliable response. More importantly, the resulting trajectory-level feedback provides contrastive evidence for memory formation. By comparing successful and failed reasoning paths, BEAM distills a concise, high-signal reasoning principle and stores it for future tasks. Thus, test-time scaling improves not only the reliability of the current prediction but also the quality and transferability of newly formed memories.

\paragraph{Case 2: Task-Conditioned Memory Utilization.}
Figure~\ref{fig:case_memory_utilization} compares memory-free reasoning with BEAM. Without memory, the executor identifies the relevant transformation but fails to resolve the structural ambiguity. BEAM instead selects memories according to their estimated task-conditioned utility and synthesizes the applicable information into actionable guidance, enabling the executor to eliminate incompatible candidates and reach the correct answer.

\end{document}